\documentclass{article}

\usepackage[final]{neurips_2021}

\usepackage[utf8]{inputenc} 
\usepackage[T1]{fontenc}    
\usepackage{hyperref}       
\usepackage{url}            
\usepackage{booktabs}       
\usepackage{amsfonts}       
\usepackage{nicefrac}       
\usepackage{microtype}      
\usepackage{xcolor}         

\usepackage{graphicx}
\usepackage{amsmath}

\DeclareMathOperator*{\argmin}{arg\,min}
\usepackage{amssymb}
\DeclareMathOperator{\pa}{\mathbf{pa}}
\DeclareMathOperator{\Pa}{\mathbf{Pa}}
\DeclareMathOperator{\doop}{\textit{do}}
\newcommand{\inner}[1]{\langle #1 \rangle}

\setcitestyle{numbers, open = {[}, close = {]}}

\title{On How AI Needs to Change to \\ Advance the Science of Drug Discovery}

\author{\textbf{Kieran Didi}\textsuperscript{\rm 1}
\quad \normalfont{and} \quad \textbf{Matej Zečević}\textsuperscript{\rm 2}\\
\textsuperscript{\rm 1}Department of Applied Mathematics and Theoretical Physics, University of Cambridge\\
\textsuperscript{\rm 2}Computer Science Department, TU Darmstadt
}

\begin{document}

\maketitle

\begin{abstract}
Research around AI for Science has seen significant success since the rise of deep learning models over the past decade, even with longstanding challenges such as protein structure prediction. However, this fast development inevitably made their flaws apparent---especially in domains of reasoning where understanding the cause-effect relationship is important. One such domain is drug discovery, in which such understanding is required to make sense of data otherwise plagued by spurious correlations. Said spuriousness only becomes worse with the ongoing trend of ever-increasing amounts of data in the life sciences and thereby restricts researchers in their ability to understand disease biology and create better therapeutics. Therefore, to advance the science of drug discovery with AI it is becoming necessary to formulate the key problems in the language of causality, which allows the explication of modelling assumptions needed for identifying true cause-effect relationships. 

In this attention paper, we present causal drug discovery as the craft of creating models that ground the process of drug discovery in causal reasoning.
\end{abstract}

\section{Limitations of Current ML Approaches to Biology}\label{sec:1}

In recent years there has been a revolution in the field of structural biology and protein engineering \cite{alquraishi}. Predicting the 3D structure of a protein has been a long standing challenge in the field, but should in principle be possible due to Anfinsen's dogma \cite{anfinsen}, according to which all information for the spatial arrangement of a protein chain is encoded in its primary structure, i.e., its amino acid sequence. The record-breaking performance of AlphaFold2 on this task of protein structure prediction from sequence clearly illustrated the power of machine learning when applied to biology\cite{jumper_highly_2021}. Many areas such as protein engineering \cite{gao} and drug discovery in general \cite{MachineLearning2019} underwent a similar major change in recent years due to the widespread adoption of deep learning based methods. While impressive advances have been made, many problems still seem out of reach.

For example, target identification, a step that occurs earlier in the drug discovery pipeline than protein engineering, relies on genomics and recently multi-omics approaches. Machine learning methods in this area are very good at identifying correlations between, for instance, the expression of a particular gene and strength of a particular disease phenotype. But many of these correlations are spurious \cite{jiang_microbiome_2019}, and may cause wrong statistical inferences and false claims of discovery in science. This problem is not limited to omics data, but is a general problem that occurs with high-dimensional data, which nowadays is present in many domains \cite{bigdata}.

In protein engineering, a recent publication showed that silent mutations are not as silent as thought, i.e., that different codons coding for the same amino acids often have significant consequences for protein fitness  \cite{shen.etal_2022}. Another paper even showed that in more flexible structural elements such as beta-sheets, the codon used has direct implications for the backbone geometry, hinting at an influence of cotranslational folding \cite{rosenberg.etal_2022}. 

According to these two papers, the DNA sequence is a very important factor in determining the structure and function of proteins since it acts as a confounder via processes such as cotranslational folding, but is so far rarely employed in algorithms used to predict protein structure and function. Here, the protein sequence is the most common input modality. Issues like this may contribute to the reproducibility problems researchers face when applying published machine learning methods to their own data \cite{reproducible}.

Since the DNA sequence corresponding to a particular protein cannot be uniquely determined from the amino acid sequence and the exact DNA sequence is in many instances not recorded, this problem can only be solved by a change in experimental design, highlighting the importance of causal understanding during the whole research process.

Further advancing the science of protein engineering will therefore require addressing these shortcomings of current machine learning approaches. A possible strategy could be to distinguish spurious from non-spurious associations to remedy problems with high-dimensional data such as in the target identification task above. 
Another approach would involve identifying hidden confounding variables that obstruct effective reasoning about the problem at hand, such as the DNA sequence in the protein engineering example above.

\begin{figure*}[t]
    \centering
    \includegraphics[width=1\textwidth]{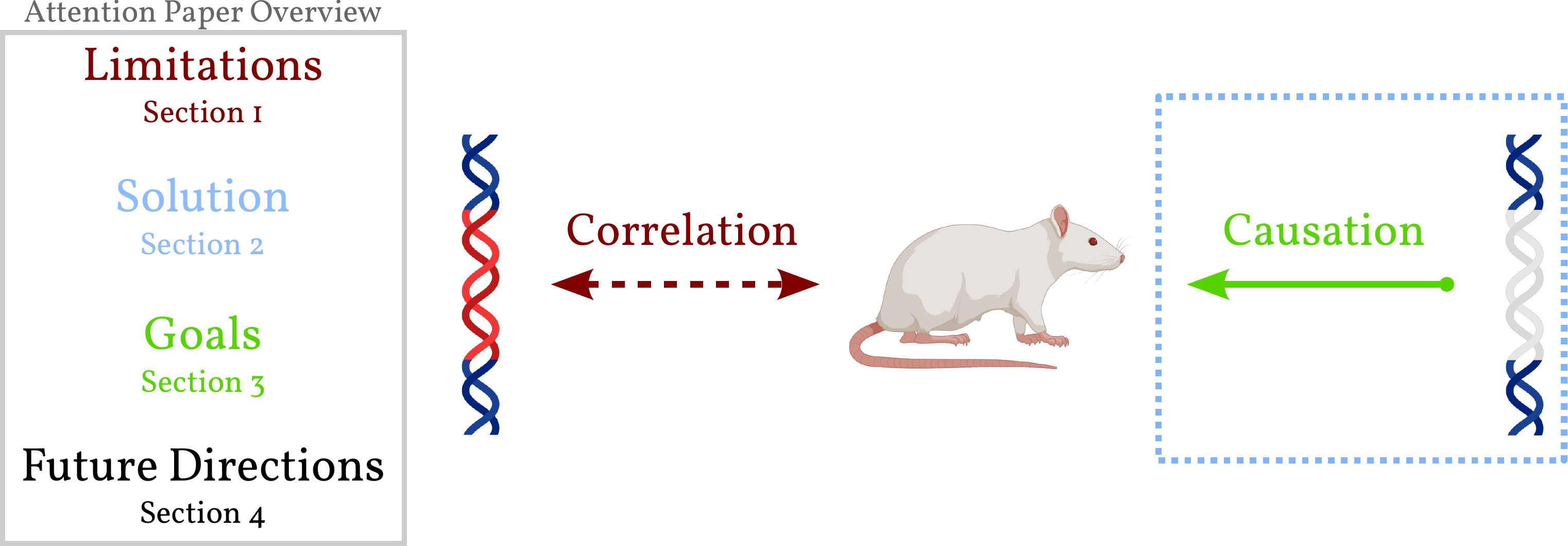}
    \caption{\textbf{Attention Paper Overview.} \label{fig:overview}
    We begin in Sec.\ref{sec:1} by discussing the limitations of current AI approaches to biology that ultimately are rooted in spurious associations between our model inputs (e.g.\ genotypes) and corresponding model outputs (e.g.\ phenotypes). Then, we propose causal reasoning methods from AI/ML as a potential solution to these limitations  in Sec.\ref{sec:2}. By using said methods, we discuss examples of how we can discover the genotype that actually causes the corresponding phenotype in Sec.\ref{sec:3}. Finally, in Sec.\ref{sec:4} we discuss future directions that can follow such discoveries to support drug discovery efforts. (Best viewed in color. Created with \url{BioRender.com})
    }
\end{figure*}

Here, we argue that incorporating insights from causal inference might shine a light on some of the issues that machine learning methods for protein engineering face and may contribute to further advancement in the field (see Fig.\ref{fig:overview}).

\section{The Study of Causation and Artificial Intelligence}\label{sec:2}
Before we discuss how causality can bridge the gap between current AI/ML practices and the science around proteins, we will build a common notion for the broad term of 'causality'. First, we make clear how science relates to causality in general (Sec.\ref{sec:21}). Then we present the formal tools with which we can specify our problems of interest (Sec.\ref{sec:22}). Finally, we consider why causality is necessary for solving the problems we care about (Sec.\ref{sec:23}).

\subsection{At the Heart of Science: Causality} \label{sec:21}
Causality can be understood as a philosophical, physical or abstract concept and studies of it date back to the ancient Greek philosopher Plato \cite{delacy1939problem}. While studies of causation generally aim at capturing its key properties precisely in a formal manner, and many attempts to this effect have been made, ironically it is something every human can grasp intuitively. To quote \cite{hernan2010causal}: ``\emph{As a human being, you are already innately familiar with causal inference’s fundamental concepts. Through sheer existence, you know what a causal effect is, understand the difference between association and causation, and you have used this knowledge consistently throughout your life. Had you not, you’d be dead.}'' Naturally, the question arises: as what is or should causality be understood? A simple answer would be that causality can be understood as the key goal of the scientific endeavour. Physicists such as Isaac Newton contributed to the scientific revolution by performing experiments that could be \emph{reproduced} and results that would be \emph{invariant} across many aspects of physical reality, such as the laws of motion, which somehow seem to depict a general `truth' about how bodies interact and move. Looking at Newton's personal notes on philosophical questions \cite{mcguire1983certain} we find: ``\emph{Plato is my friend, Aristotle is my friend, but my greatest friend is truth.}'' We might conclude from these observations that at the intersection of science and causation we find `truth'. Science seeks truth, since one of its implications would be to better understand physical reality. Causality on the other hand can provide that truth since it will know which relations are only spurious and which are indeed consequences of each other. We can also find this duality of science and causation in any typical experimental procedure that we perform as scientists. For instance, we might wish to infer whether the rooster's crow causes the sun to rise. To do so, we conduct a scientific experiment in which we place the rooster in a soundproof box, that is, we change one aspect of the environment (namely silencing the rooster's crow) while keeping everything else the same. At the end of the experiment it turns out to that the sun still rose, leading to the conclusion that the rooster's crow and the sunrise are independent. With our scientific experiment, which included an intervention, we ultimately performed causal inference. Therefore, we might even understand \emph{science as the pursuit of causality}.

\subsection{A Formalization of Causality} \label{sec:22}
Arguably the greatest success thus far in the formalization of causality is attributed to the Pearlian counterfactual theory of causation \citep{pearl2009causality}. Several works in cognitive science have even been in support of that theory as a candidate formalization for how humans conduct their mental inferences \cite{gerstenberg2017eye,lagnado2013causal}. Following said Pearlian notion, a Structural Causal Model (SCM) is defined as a 4-tuple $\mathcal{M}:=\inner{\mathbf{V},\mathbf{U},\mathcal{F},P(\mathbf{U})}$ where the so-called structural equations
\begin{align}\label{eq:scm}
v_i \leftarrow f_i(\pa_i,u_i) \in \mathcal{F}
\end{align} assign values (denoted by lowercase letters) to the respective endogenous/system variables $V_i\in\mathbf{V}$ based on the values of their parents $\Pa_i\subseteq \mathbf{V}\setminus V_i$ and the values of some exogenous variables $\mathbf{U}_i\subseteq \mathbf{U}$ (sometimes also referred to as unmodelled or nature terms), and $P(\mathbf{U})$ denotes the probability function defined over $\mathbf{U}$. The SCM model class comes with several interesting properties. The model class induces a causal graph $G$, further it induces an observational/associational distribution over $\mathbf{V}$ (typical question ``What is?'', example ``What do the symptoms tell us about the disease?''), and it can generate infinitely many interventional/hypothetical distributions (typical question ``What if?'', example ``What if I take an aspirin, will my headache be cured?'') and counterfactual/retrospective distributions (typical question ``Why?'', example ``Was it the aspirin that cured my headache?'') by using the $\doop$-operator which ``overwrites'' structural equations.
An intervention on an SCM occurs when (multiple) structural equations are replaced through new non-parametric functions, thus effectively creating an alternate SCM. Interventions are referred to as \emph{imperfect} if the parental relation is kept intact, as \emph{perfect} if not, and even \emph{atomic} when the intervened values are also kept constant. It is important to realize that interventions are of fundamentally \emph{local} nature, and the structural equations (variables and their causes) dictate this locality. This further suggests that mechanisms remain invariant to changes in other mechanisms. An important consequence of said autonomic principles is the \emph{truncated factorization}
\begin{align} \label{eq:truncatedfactorization}
p^{\mathcal{M}_{do(\mathbf{w})}}(\mathbf{v}) = p^{\mathcal{M}}(\mathbf{v}_{\mathbf{w}}) = \prod\nolimits_{V_i\notin \mathbf{W}} p^{\mathcal{M}}(v_i\mid \pa_i)
\end{align}
derived by \cite{pearl2009causality}, which gives the probability of observing variables $\mathbf{V}$ to be of value $\mathbf{v}$ if variables $\mathbf{W}$ had been $\mathbf{w}$\footnote{In the literature, to distinguish this \emph{assignment} (or alternate world) $\mathbf{W}=\mathbf{w}$ from a pure conditional $p(\mathbf{v}\mid \mathbf{w})$ the $\doop$-notation is often employed, that is, $p(\mathbf{v}_\mathbf{w})=p(\mathbf{v}\mid \doop(\mathbf{w}))\neq p(\mathbf{v}\mid \mathbf{w})$.}. We also see that an intervention introduces an independence between the set of nodes intervened with and their causal parents. For notational convenience, we use $p^{\mathcal{M}}$ and $p$ interchangeably if clear from context. The definition of $\mathcal{M}$ also allows for shared $\mathbf{U}$ terms between the different $V_i$, which is what is known as a \emph{hidden confounder} since it is a common cause of at least two $V_i,V_j (i\neq j)$. In contrast, a ``common" confounder would be a common cause from within $\mathbf{V}$. In this sense, the SCM extends the famous model class of Bayesian Networks \cite{pearl2011bayesian} by allowing for both reasoning about counterfactuals and hidden confounders \cite{bongers2021foundations}. The SCM's applicability to AI/ML has been illustrated successfully in various applications involving fairness \cite{kusner2017counterfactual}, healthcare \cite{bica2020time}, marketing \cite{hair2021data} and education \cite{hoiles2016bounded}. As suggested by the Causal Hierarchy Theorem (CHT; \cite{bareinboim20201on}), the properties of an SCM form the Pearl Causal Hierarchy (PCH) consisting of different levels of distributions, namely $\mathcal{L}_1$ \emph{associational}, $\mathcal{L}_2$ \emph{interventional} and $\mathcal{L}_3$ \emph{counterfactual}. The PCH suggests that causal quantities ($\mathcal{L}_i,i\in\{2,3\}$) are in fact richer in information than statistical quantities ($\mathcal{L}_1$), and that there exists a necessity of causal information (e.g.\ structural knowledge, essentially ``outside'' model knowledge) for inference based on lower rungs. Put differently, $\mathcal{L}_1 \not\Rightarrow \mathcal{L}_2$, therefore to reason about $\mathcal{L}_2$ or to \emph{identify} such causal quantities we need more than only observational data from $\mathcal{L}_1$. These levels or languages (such as in logic) differ in that $\mathcal{L}_1$ is ``common'' statistics (any quantity of type $p(\mathbf{A})$\footnote{Note we can always define $\mathbf{A}:=\mathbf{B}\mid\mathbf{C}$ to cover conditionals.} derivable by the laws of probability theory and Bayes rule), $\mathcal{L}_2$ are expressed through interventions $p(\mathbf{A}_{\mathbf{b}})$ and $\mathcal{L}_3$ are conjunctions of the former $p(\mathbf{A}_{\mathbf{b}},\dots,\mathbf{C}_{\mathbf{b}})$. To evaluate any of the levels, we can use the following formula,
\begin{equation} \label{eq:l12val}
p(\mathbf{a}_{\mathbf{b}},\dots,\mathbf{c}_{\mathbf{d}}) = \sum_{\mathcal{U}} p(\mathbf{u}) \quad \text{where} \quad \mathcal{U}=\{\mathbf{u}\mid \mathbf{A}_{\mathbf{b}}(\mathbf{u})=\mathbf{a},\dots, \mathbf{C}_{\mathbf{d}}(\mathbf{u})=\mathbf{c}\},
\end{equation} for instantiations $\mathbf{a},\mathbf{b},\mathbf{c},\mathbf{d}$ of the node sets $\mathbf{A},\mathbf{B},\mathbf{C},\mathbf{D} \subseteq \mathbf{V}$ which represent different ``worlds''. The $\mathbf{A}_{\mathbf{b}(\cdot)}$ functions take the $\mathbf{U}$ as argument since they are the only source of uncertainty in an SCM, as the structural equations are \emph{deterministic}. Eq.\ref{eq:l12val} captures all the rungs of the PCH, for $\mathcal{L}_1$ we might only consider $\mathbf{A}=\mathbf{A}_{\emptyset}$ whereas for $\mathcal{L}_2$ a single alternate world $\mathbf{A}_{\mathbf{b}}$. Put in words, the sum of the latent aspects of our SCM that agree on the given specification (the elements of $\mathcal{U}$) captures perfectly our belief in that specification given the assumptions.
\begin{figure*}[t]
    \centering
    \includegraphics[width=1\textwidth]{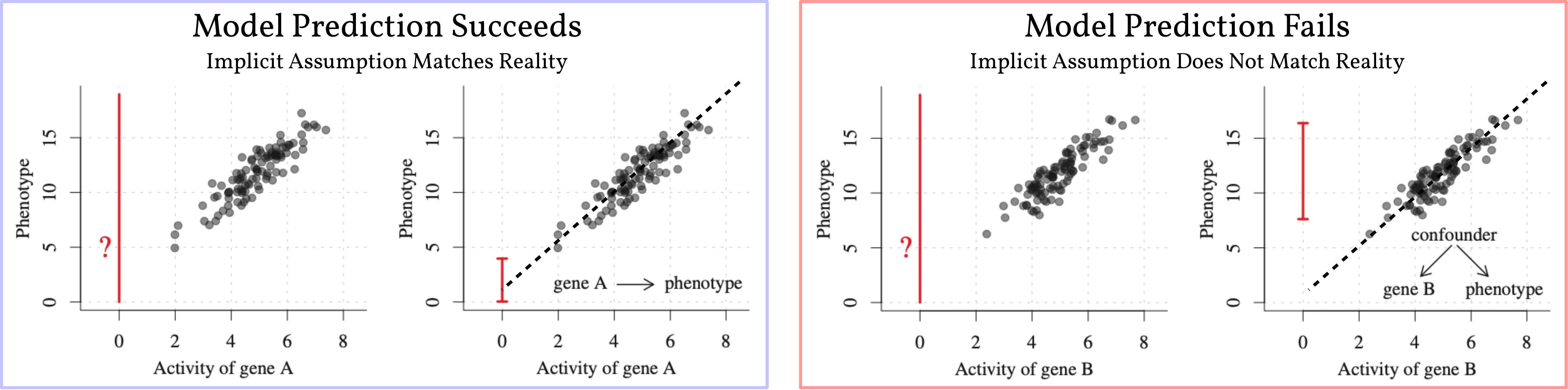}
    \caption{\textbf{Necessity of Causality for AI for Science.} \label{fig:necessity}
    While in some instances predictive algorithms that just use correlation-based insights might succeed in their prediction (left), the assumption that correlation implies causality does not need to hold and can cause these kind of algorithms to fail (right). Refer to Section \ref{sec:23} for a detailed description. Modified and inspired by \cite{peters2017elements}. (Best viewed in color.)
    }
\end{figure*}

\subsection{On Why AI for Science is Hopeless Without Causality} \label{sec:23}
The recent strides in AI for Science might be solely characterized by learning methods, practically becoming synonymous with ML. However, ML is concerned with data and learning and arguably not so much with ``what lies outside the data'' (that is, modelling assumptions). In the previous sections, we have seen how causality is the discipline of capturing modelling assumptions precisely and how we can phrase it in a probabilistic manner that is compliant with modern AI/ML. However, is causality really necessary for AI for Science? With the impressive achievements in the area over recent times it becomes ever more difficult to be a `non-believer' of the \emph{scaling hypothesis} \citep{gwern2020,sutton2019bitter} especially with the advent of large scale models in the AI/ML community. The scaling hypothesis captures the idea of emergent properties as a result of scaling neural network architectures in terms of parameters and data (rooting parts of the overarching idea in results from neuroscience that suggest the human brain is `just' a scaled up primate brain \citep{herculano2012remarkable}). However, the CHT (see previous section) does \emph{prove} the impossibility of PCH cross-layer inference \emph{without modelling assumptions}. We can even illustrate the argument with a rather simple example taken from \cite{peters2017elements}, see Fig.\ref{fig:necessity}. 

We observe two different data sets: one relates the activity of a certain gene which we call A to a phenotype (for different measurements of gene A and the phenotype) and the other of a gene B to a phenotype. We can visualize the data using scatter plots. We see a strong correlation, which is similar in both plots. However, what would \emph{hypothetically} speaking happen if we were to delete any of the genes? It turns out the predictions would be strongly different since the underlying causal model is different. Gene A directly causes the phenotype, whereas for gene B it is simply confounded with the phenotype. Therefore, had we deployed a simple regression model, we would have only answered correctly in the former case. Thus, we conclude that if we do not consider causal modelling assumptions outside the data, then AI for Science becomes hopeless as the best answer that the model could give would be ``don't know.''

\section{Using Causal AI to Advance Drug Discovery} \label{sec:3}

\begin{figure*}[t]
    \centering
    \includegraphics[width=1\textwidth]{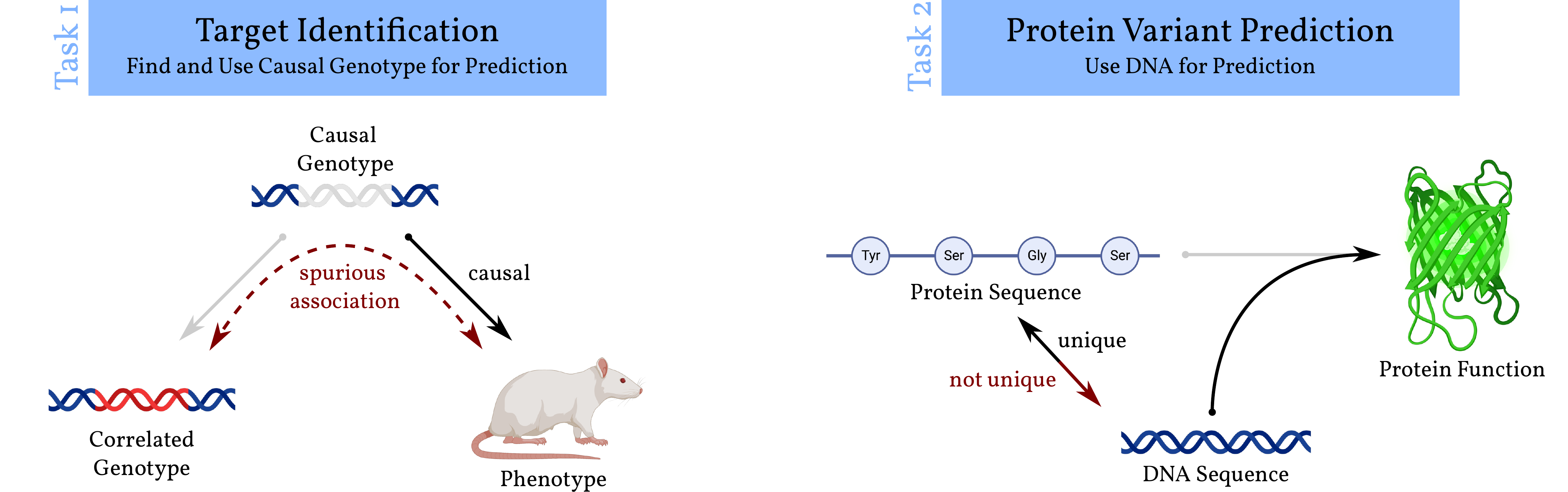}
    \caption{\textbf{Target Identification and Protein Variant Prediction.} \label{fig:tasks}
    Two examples on how causal approaches can help in drug discovery. Left: in target identification, we often have several correlated genotypes for one phenotype, but causal inference can help us identify the true causal genotype which we need to perturb in order to affect the phenotype. Right: protein sequences are the typical input for protein function prediction, but using DNA sequences instead could capture causal dependencies that cannot be inferred from the protein sequence alone, such as cotranslational folding effects. Refer to Section \ref{sec:3} for a detailed description of both examples. (Best viewed in color. Created with \url{BioRender.com})
    }
\end{figure*}

So far, we highlighted some problems of current AI approaches to biology and introduced the notion of causality as a possible solution. In this section, we want to illustrate some goals that may be achieved by formulating research problems from biology (more specifically, drug discovery) in the language of causality. Both tasks that we formalize are illustrated in Fig.\ref{fig:tasks}.

\textbf{Task 1: Target Identification}

Target identification is one of the earlier steps in the drug discovery pipeline in which we want to identify the cause of a disease phenotype that we want to target (for example, a particular gene). In a typical set-up for this target identification process, we would take gene expression data as input and some phenotype as output (for example \emph{oncogenesis}, the formation of cancer). The goal is to find the gene that is actually causing the abnormal phenotype (increased oncogenesis) in the flood of data typically resulting from such a gene expression experiment.


Formally, let $\mathbf{V}_G$ be the set of different genes that we measure and $\mathbf{X}\in\mathbb{R}^{|\mathbf{V}|\times|\mathbf{M}|}$ the data matrix in which we collect our experimental data, with each row corresponding to expression data ($\mathbf{M}$) for one particular gene (if we measure gene expression via a single numerical value, $|\mathbf{M}| = 1$). We also measure a target vector that describes the phenotypical change, denoted $\mathbf{y}\in\mathbb{R}^{|\mathbf{V}|}$, for phenotype $V_P$. The assumption we are willing to make is that $\mathbf{V}_G$ will contain the set of genes that have a causal effect on $V_P$, that is, there exists a structural equation $f$ and some set $\Pa_P \subseteq \mathbf{V}_G$ such that $V_P = f(\Pa_P, U_P)$ for some $U_P$. In terms of the formalism introduced in Sec.\ref{sec:22}, we are therefore interested in discovering $\Pa_P$ by leveraging our data $(\mathbf{X},\mathbf{y})$. Note that it is not necessary to uncover the underlying causal graph but only the causal parents to $V_P$. Our objective can be written as
\begin{equation}
    \mathbf{W}^* = \argmin_{\mathbf{W}\subseteq \mathbf{V}_G} \mathcal{L}(\mathbf{W}; \mathbf{X},\mathbf{y})
\end{equation} where the loss function $\mathcal{L}$ should be designed such that the global optimum will correspond to finding the true parents, $\mathbf{W}^*=\Pa_P$. Typically, in AI/ML, $\mathcal{L}$ will simply be an error measurement in prediction, which therefore cannot guarantee that the desired equality will hold. Also, for \emph{general} SCM with no further assumptions it has been shown to be impossible \cite{peters2017elements}. Nonetheless, existing works such as Invariant Causal Prediction \cite{peters2016causal} can leverage knowledge on for instance different environments (which in this case could be the same experiment conducted in different labs under different instruments and protocols) to guarantee the desired equality.

By identifying the true causal parents $\mathbf{W}^*$, we are able to make sense of the experimental data and receive guidance on what to do next in our drug discovery effort (for example, knock out this particular gene to test our causal hypothesis).




\textbf{Task 2: Protein Variant Prediction}


Protein engineering can be seen as a later stage in the drug discovery pipeline for biomolecules, but has applications far beyond the development of drugs. In this field, we often have a specific protein sequence coding for a protein of interest and want to make changes to this protein sequence in order to engineer the protein function in a way that suits our interests (for example, to bind to a specific molecule or catalyze a specific reaction). To facilitate this process, we would like an algorithm that performs \emph{Protein Variant Prediction} i.e., that tells us what effect a specific mutation will have \emph{before} we actually perform this intervention.

Formally, let $\mathbf{V}_A$ denote the set of different amino acids that make up a protein sequence. We are given several protein sequences, that is `codewords', consisting of ordered tuples $\mathbf{p}=(a_1,\dots,a_d)$ where $a_i\in\mathbf{V}_A$. Further, for every protein sequence $\mathbf{p}$ we observe a mutated sequence $\mathbf{p}'$, with modified amino acids $\mathbf{m}$, such that $\mathbf{p}_{\setminus \mathbf{m}} \subset \mathbf{p}'$. These sequences are collected in a data matrix $\mathbf{X}\in\mathbb{R}^{(m+1)\times d}$ where $m$ is the number of mutated protein sequences. We also measure the protein function $Y$ for said interventions on protein sequences denoted $\mathbf{y}\in\mathbb{R}^{m+1}$. Our goal is to quantify how the mutation $\mathbf{m}$ affects the measured protein function. In terms of the formalism introduced in Sec.\ref{sec:22}, we are interested in the causal effect of $\mathbf{p}$ on $\mathbf{y}$. That is, we want to uncover the structural equation that governs the relation $\mathbf{y}=f_Y(\mathbf{p}, U_Y)$\footnote{If we choose not to explicitly capture $U_Y$, we can use $p(\mathbf{y}\mid \doop(\mathbf{p}))$ as a proxy for $f_Y$.}. Since $\mathbf{X}$ captures all the information on the experimental procedure for going from $\mathbf{p}$ to $\mathbf{p}'$ via $\mathbf{m}$, we can optimize the following objective
\begin{equation}\label{eq:t2}
    f^* = \argmin_{f\in\mathcal{F}} \mathcal{L}(f; \mathbf{X},\mathbf{y})
\end{equation} where $\mathcal{F}$ denotes the space of all possible structural equation parameterizations\footnote{Typically, in current AI/ML research we would choose neural network architectures as our hypothesis class.} under consideration and ideally we have $f^*=f_Y$. Special cases of Eq.\ref{eq:t2} include the estimation of average treatment effects for binary variable systems, for which several off-the-shelf computational libraries such as \cite{sharma2020dowhy,chen2020causalml} are readily available. 

It is important to note the underlying assumption for the formulation in Eq.\ref{eq:t2}: the information in $\mathbf{X}$ should provide sufficient information to compute $\mathbf{y}$ using $f_Y$. While current approaches typically use the protein sequence as input, recent publications show that this might not be enough. It was shown that additional causal effects such as silent mutations which affect the protein function are captured by the DNA sequence and not by the protein sequence \cite{shen.etal_2022,rosenberg.etal_2022}, suggesting that the underlying assumption might not hold. Our causal approach to this problem explicates the issue in a formal language and offers a route to solve it by including the true causal parent of protein function (DNA sequence) instead of the protein sequence.





\section{Future Directions: Putting our Causal Understanding on Trial}\label{sec:4}

Biological research in its current form is very focused on individual components and interactions\footnote{This may be due to the inherent complexity of biological systems coupled with the recent advances in high-throughput technologies.} and thereby makes drawing conclusions about the whole system very difficult, if not impossible \cite{lazebnik_can_2002}. 

For the study of emergent properties omnipresent in biology, a more sophisticated approach to inference is needed. This is valid for both conceptually thinking about problems, as illustrated earlier in the case of protein engineering, and knowing in which cases AI/ML can help scientists gain actionable insights from the ever-increasing amounts of data they face \cite{bizzarri_call_2019}.

There is a growing awareness that causal approaches to AI/ML can help solve challenges in biology, but this has so far been limited to certain subfields such as network biology \cite{lecca_machine_2022}. In this attention paper we argued that many stages of the drug discovery pipeline might benefit from incorporating the recent advances in causal reasoning and therefore support scientists in their quest to design new therapeutics. Specifically, we provided a formalization for \emph{Target Identification} (Task 1) and \emph{Protein Variant Prediction} (Task 2) in terms of causality. We hope that the provided formalizations can serve as a blueprint to (I) how researchers can use the Pearlian notion to causality to explicate modelling assumptions but also (II) how researchers can use causal AI/ML to solve scientific questions such as those found in drug discovery.

\textbf{Remark on Societal \& Ethical Implications.} Drug discovery is essential for a healthy, thriving society. Improvements in AI for the science of drug discovery, via causal AI/ML, could directly translate into both cost reduction in the monetary sense, e.g., that less experiments are required in total since we can discriminate non-/spurious associations, as well as a societal sense e.g., more people can be helped with corresponding therapeutics.

\textbf{Acknowledgments.} The authors acknowledge the support of the German Science Foundation (DFG) project “Causality, Argumentation, and Machine Learning” (CAML2, KE 1686/3-2) of the SPP 1999 “Robust Argumentation Machines” (RATIO). This work was supported by the ICT-48 Network of AI Research Excellence Center “TAILOR” (EU Horizon 2020, GA No 952215), the Nexplore Collaboration Lab “AI in Construction” (AICO) and by the Federal Ministry of Education and Research (BMBF; project “PlexPlain”, FKZ 01IS19081). It benefited from the Hessian research priority programme LOEWE within the project WhiteBox \& the HMWK cluster project “The Third Wave of AI” (3AI).

\bibliography{main}


\end{document}